%% file: shamsi2026_ridge_selective_prediction.tex
\documentclass[conference]{IEEEtran}
\IEEEoverridecommandlockouts

\usepackage{cite}
\usepackage{amsmath,amssymb,amsfonts,bm}
\usepackage{graphicx}
\usepackage{textcomp}
\usepackage{xcolor}
\usepackage{booktabs}
\usepackage{array}
\usepackage{subcaption}
\usepackage{tikz}
\usepackage{pgfplots}
\pgfplotsset{compat=1.18}
\usetikzlibrary{arrows.meta,calc,positioning,decorations.markings}
\usepackage[subpreambles=true]{standalone}
\usepackage{fancyhdr}
\usepackage{etoolbox}

\graphicspath{{../}{../figures/}}

\newcommand{\bbR}{\mathbb{R}}
\newcommand{\bbE}{\mathbb{E}}
\newcommand{\bbP}{\mathbb{P}}
\newcommand{\calR}{\mathcal{R}}
\newcommand{\calX}{\mathcal{X}}
\newcommand{\calV}{\mathcal{V}}
\newcommand{\calQ}{\mathcal{Q}}
\newcommand{\bH}{\mathbf{H}}
\newcommand{\bh}{\mathbf{h}}
\newcommand{\bx}{\mathbf{x}}
\newcommand{\bv}{\mathbf{v}}
\newcommand{\bg}{\mathbf{g}}
\newcommand{\bV}{\mathbf{V}}
\newcommand{\bLam}{\boldsymbol{\Lambda}}
\newcommand{\bmu}{\boldsymbol{\mu}}
\newcommand{\bsigma}{\boldsymbol{\sigma}}
\newcommand{\tbx}{\tilde{\mathbf{x}}}

\begin{document}

\fancypagestyle{extabs}{%
  \fancyhf{}%
  \renewcommand{\headrulewidth}{0pt}%
  \fancyhead[L]{\footnotesize\itshape Extended Abstract}%
}
\pagestyle{extabs}
\AtBeginDocument{\thispagestyle{extabs}}
\makeatletter
\patchcmd{\maketitle}{\thispagestyle{plain}}{\thispagestyle{extabs}}{}{}
\makeatother

\title{Density Ridge Selective Prediction \\for LLM and VLM Hallucination Detection \\under Calibration-Label Scarcity}

\author{\IEEEauthorblockN{Nina I. Shamsi}
\IEEEauthorblockA{\textit{Northeastern University}\\
Boston, United States \\
nshamsi@ece.neu.edu}
}

\maketitle

\begin{abstract}
Hallucination detection in large language and vision-language models is
increasingly framed as selective prediction, where a detector assigns a
confidence score and abstains when confidence is low. Unsupervised sampling
detectors (Semantic Entropy) avoid labels but plateau in quality,
while supervised probes attain stronger in-distribution scores yet
degrade sharply when calibration labels are scarce. We recover the response
manifold of an LLM as the density ridge of a kernel density estimate built on
a six-dimensional kinematic feature map of hidden state generation
trajectories. A test generation is scored by the negated Euclidean distance
from its projected feature point to the nearest ridge vertex, yielding a
low-dimensional geometric skeleton of the stochastic output distribution. We
evaluate against Semantic Entropy, topological methods, and
log-probability on six QA benchmarks (HaluEval-QA, TriviaQA,
GSM8K, POPE, ScienceQA, A-OKVQA) using eight text and vision LLMs in a
deliberately label-scarce protocol ($n_{\text{cal}}{=}200$ queries,
$N{=}5$ generations). Our ridge-based score beats on AUROC with 5-20 points gain, while
demonstrating tempered degradation under calibration-label scarcity.
\end{abstract}

\begin{IEEEkeywords}
hallucination detection, selective prediction, density ridge, kernel density
estimation, large language models, uncertainty quantification.
\end{IEEEkeywords}

\section{Introduction}
Selective prediction abstains when a confidence score falls below a
threshold. Two dominant families dominate the literature on LLM hallucination
detection: sampling-based unsupervised detectors that probe output dispersion
\cite{kuhn2023,chenLLMsInternalStates2024}, and supervised hidden state probes
that train a classifier on labeled correctness annotations. The latter
outperform the former under abundant labels but suffer in deployments where
calibration data is scarce \cite{kossen2024}. We exploit a complementary
signal: the geometry of generation-time hidden state trajectories. Repeated
sampling at a fixed query traces a multimodal distribution in embedding space
whose modes encode distinct response strategies. Recent work on the curvature
evolution of LLM trajectories has shown such kinematics to be diagnostic of
reasoning quality \cite{trace2025,tracedet2025}. We characterize this
distribution by the \emph{density ridge}~\cite{ozertem2011,genovese2014} of a
KDE fitted to kinematic features of correct trajectories, and score test
generations by proximity to this 1-D ridge.

\paragraph*{Contributions.}
(i) A response-manifold detector recovering the LLM response manifold as a
density ridge over supervised, low-dimensional trajectory-curvature features.
(ii) A label-scarce evaluation across nine models and seven benchmarks
against Semantic Entropy, log-probability, and topology  
baselines. (iii) Ablations isolating the contribution of ridge geometry
across three parameterization variants and kernel configurations.

\begin{figure*}[t]
  \centering
  \input{figures/tkde.tex}
  \caption{Trajectory-branch SCMS confidence pipeline.
  \textbf{(a)} For each query $q_i$ and each of $N$ generations, the
  hidden-state trajectory $\bH_{i,j}\!\in\!\bbR^{T_{i,j}\times D}$ is colored
  by the binary correctness label $y_i$. \textbf{(b)} The feature map
  $\varphi(\bH)=(K_n,R^2,\dots)$ projects each trajectory to a point
  $\bx_m\in\bbR^6$, producing $\calX_{\text{traj}}$ (two coordinates shown).
  \textbf{(c)} The KDE $\hat p_h$ over the trusted subset
  $\calX_{\text{traj}}^{+}=\{\bx_m:y_{\sigma(m)}=1\}$ is run through SCMS to
  extract the raw ridge. \textbf{(d)} Intrinsic-dimension verification clamps
  $r{=}1$, yielding the 1-D ridge $\calR_1$ sampled as vertices
  $\calV=\{\bv_k\}_{k=1}^{K}$. \textbf{(e)} A test query $q_\star$ contributes
  trajectory points $\bx_{\star,j}$; the score is the negated mean
  perpendicular distance to the nearest ridge vertex.}
  \label{fig:traj-pipeline}
\end{figure*}
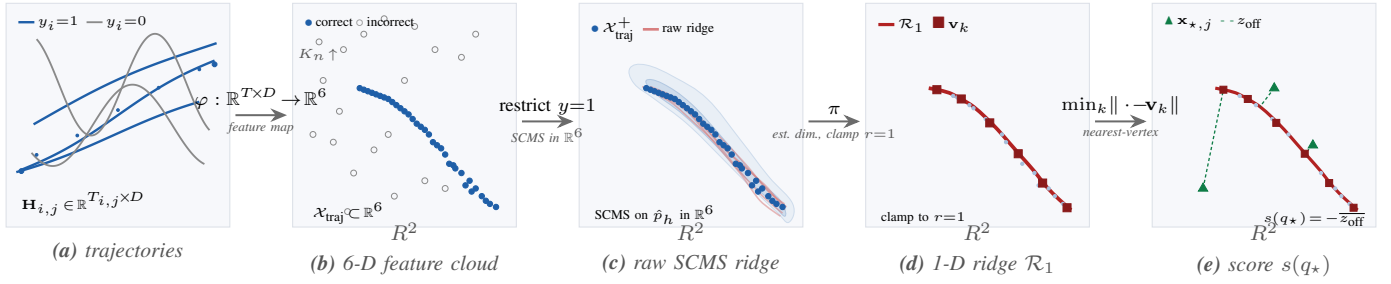

\section{Method}
\label{sec:method}

\textbf{Kinematic feature map.} Given training queries
$\calQ_{\text{train}}=\{q_i\}$ with correctness labels $y_i\in\{0,1\}$, each
sampled completion $j$ induces a hidden-state trajectory
$\bH_{i,j}\in\bbR^{T_{i,j}\times D}$ of final-layer, last-token states. With
$\Delta\bh_t=\bh_{t+1}-\bh_t$ and discrete curvature
$\kappa_t=\|\Delta\bh_{t+1}-\Delta\bh_t\|_2/\bigl[\tfrac12(\|\Delta\bh_t\|_2+\|\Delta\bh_{t+1}\|_2)\bigr]^2$,
the feature map $\varphi:\bbR^{T\times D}\to\bbR^6$ collects
$(K_n,R^2,K_{\max},\overline{\|\Delta\bh\|},\|\bv\|_{\max},\tau^\star)$:
mean and peak curvature, displacement linearity, mean and peak per-step
displacement, and the normalized argmax of $\kappa_t$
\cite{trace2025}. Each generation becomes $\bx_m=\varphi(\bH_m)\in\bbR^6$,
$z$-scored by training statistics
$\bmu_{\text{train}},\bsigma_{\text{train}}$.

\textbf{Ridge construction.} On the correct subset
$\tilde\calX^+=\{\tbx_m:y_{\sigma(m)}{=}1\}$ we fit a Gaussian KDE
$\hat p$ with bandwidth via Scott's rule in $d{=}6$. With
$\bg=\nabla\log\hat p$ and Hessian
$\nabla^2\log\hat p=\bV\bLam\bV^\top$
($\lambda_1\le\dots\le\lambda_6$), the 1-D density
ridge~\cite{ozertem2011,genovese2014} is
$\calR_1=\{\tbx:\bV_\perp^\top\bg=\mathbf{0},\,\lambda_5<0\}$, with
$\bV_\perp\in\bbR^{6\times5}$ spanning the normal subspace. SCMS iterates
the projected gradient $\bV_\perp\bV_\perp^\top\bg$ to fixed points
$\calV=\{\bv_k\}$. Intrinsic-dimension verification via TwoNN
\cite{facco2017} supports the $r{=}1$ clamp. Three chart variants
(Table~\ref{tab:ridgevariants}) furnish usable coordinates: global LTSA, the
Hastie--Stuetzle geodesic arclength \cite{hastie1989}, and a stitched
local-chart atlas \cite{roweis2002,brand2003}.

\input{tables/ridge_variants}

\textbf{Score and OOD interpretation.} For a test query with trajectories
$\{\bH_{\star,j}\}$, the off-ridge distance is
$z_{\text{off}}(\tbx_{\star,j})=\min_k\|\tbx_{\star,j}-\bv_k\|_2$ and the
score $s(q_\star)=-\tfrac{1}{N_\star}\sum_j z_{\text{off}}(\tbx_{\star,j})$.
Because $\calR_1$ is fit on $\nu^+$ alone, incorrect points are OOD
regardless of their own density. Standard regularity \cite{chen2015} gives
$d_H(\hat\calR_1,\calR_1)=O(h^2)+O_\bbP(\sqrt{\log n/(nh^8)})$ and an
expected score gap
$\bbE_{\nu^-}[z_{\text{off}}]-\bbE_{\nu^+}[z_{\text{off}}]\ge
\Delta-\rho^+-O(h^2)$ concentrating at rate $N_\star^{-1/2}$.

\section{Experiments}
\label{sec:exp}

{\makeatletter
\renewenvironment{table}{\@dblfloat{table}}{\end@dblfloat}
\makeatother
\input{tables/selective_pred_table}}

{\makeatletter
\renewenvironment{figure}{\@dblfloat{figure}}{\end@dblfloat}
\makeatother
\input{figures/metrics}}

\subsection{Setup}
We evaluate eight text and vision-language models, including
Mistral-7B-Instruct-v0.3, Gemma-2-9B-IT, LLaVA-1.5-7B, Idefics3-8B-Llama3,
and SmolVLM-Instruct, on six QA benchmarks spanning textual factuality
(HaluEval-QA, TriviaQA), mathematical reasoning (GSM8K), and
multimodal grounding (POPE, ScienceQA, A-OKVQA). A \emph{cell} denotes one
(model, dataset, quantification) combination. To simulate data scarce deployments, queries are
retained only if they yield at least $\ge 3-6$ correct generations, and calibration
uses $n_{\text{cal}}=200$ queries with $N=5$ generations each
(test size is 60). We report AUROC and PRR (higher is better), and
AURC and AUGRC (lower is better). The head-to-head comparison is restricted
to cells common to all detectors.

\subsection{Baselines}
We compare against unsupervised sampling detectors Semantic Entropy
\cite{kuhn2023}, log-probability, and topology-based metrics such as topological LID-MLE and persistent-homology, the
TRACED mean curvature scalar \cite{trace2025}, and naive embedding-geometry
baselines ($k$NN-$R^2$, PCA-based, flow matching to the correct-class
centroid). Recent representation-manifold approaches have been explored
elsewhere for safety \cite{kanMANATEEInferenceTimeLightweight2026}; the
multi-dimensional nature of LLM features \cite{engelsNotAllLanguage2025}
motivates the 6-D kinematic descriptor over scalar summaries. The ordering of detectors by AUROC is: ridge $>$ log-probability $>$
single-scalar trajectory summaries $>$ naive geometry $>$ topological
summaries. Negative-control scalars (initial-state distance, weight-norm)
and maximum-token log-probability invert to PRR $<0$, as anti-correlated
signals should.

\subsection{Ablations}
\paragraph{Kernel configuration.}
Sweeping eleven SCMS kernel variants together with three naive-geometry
baselines (PC1, $k$NN, Mahalanobis) within each cell, the canonical kernel
(fixed bandwidth via Scott's rule, uniform weights, sample covariance,
$r{=}1$ constraint) attains the best mean rank (3.27), and every SCMS
variant outranks all three naive baselines (mean ranks $\ge 11$). The
ridge structure, not generic embedding distance, is what separates
correct from hallucinated generations.

\subsection{Main Results}
Table~\ref{tab:selective_pred_combined} reports per-cell selective
prediction metrics for three representative detector classes: the ridge
score (Ridge Arclength, Ridge LTSA Chart, or Ridge Atlas), a non-ridge
geometric or topological baseline (Topology LID-MLE, $k$NN-$R^2$, Flow
Matching, Semantic Entropy, or TRACED mean curvature), and calibrated
sequence log-probability. The ridge score is the most performant predictor
on every metric for all cells.

\textbf{Quantification trials across detectors and datasets.}
The log-probability baseline is the strongest non-ridge competitor on the
textual factuality and grounding splits: it attains AUROC 0.913 on
Mistral/HaluEval-QA, 0.817 on LLaVA/POPE, and 0.79--0.80 on the SmolVLM
and Idefics3 multimodal cells. Yet on every cell the ridge variant
improves AUROC by 5--20 points absolute and concurrently reduces AURC and
AUGRC. The largest gains occur on the multimodal grounding datasets
(A-OKVQA, ScienceQA), where Idefics3 sees AUROC rise from 0.795 (logP) to
0.972 (Ridge LTSA Chart) and SmolVLM/ScienceQA rises from 0.790 to 0.990
(Ridge Arclength), with AUGRC roughly halved. The non-ridge baselines are
markedly less stable: Topology LID-MLE degrades to near-chance (AUROC
0.470--0.511) on SmolVLM/A-OKVQA and Idefics3/ScienceQA, while Flow
Matching to the correct-class centroid (AUROC 0.769 on SmolVLM/ScienceQA)
and Semantic Entropy (0.740 on LLaVA/POPE) underperform log-probability
on the cells where they are the strongest non-ridge entrant. On the
reasoning-heavy text cells where log-probability is already strong (HaluEval-QA), the
ridge variant still recovers a 4--6 point AUROC margin while trimming
AURC by roughly 15\%, indicating that the geometric signal is
complementary to, not redundant with, token-level confidence.
Figure~\ref{fig:metrics-top3} aggregates the top-3 detectors per cell
under bf16 (top) and nf4 (bottom) precision, illustrating that the
ridge--logP--non-ridge ordering persists under aggressive quantization.

\section{Discussion and Conclusion}
That the ridge detector exceeds both unsupervised sampling detectors and
sequence log-probability under deliberately scarce calibration labels
indicates that what separates faithful from hallucinated generations is the
\emph{shape} of the trajectory-feature space, not its raw location. Every
SCMS variant outranks PC1, $k$NN, and Mahalanobis: generic embedding
distance is insufficient, and the recovered 1-D manifold density ridge is what is utilized by the score. \textbf{Limitations.} (i) The head-to-head
comparison is restricted to the cells common to all detectors. (ii)
Semantic Entropy is at chance on closed-form slices, which is an expected
degeneracy under data and label scarcity. (iii) The supervised projection front-end requires both-class
labels at fit time. Natural extensions include genuine per-axis kernel
ablations and lifting distribution-free conformal guarantees
\cite{angelopoulos2022crc,bates2023sscb} from a companion probe onto the
ridge score itself.

\end{document}

%% file: figures/tkde.tex
%
%
%



{%
\definecolor{TPcorrect}  {HTML}{1F5FA8}
\definecolor{TPincorrect}{HTML}{8A8A8A}
\definecolor{TPridge}    {HTML}{B22222}
\definecolor{TPridgeV}   {HTML}{8A1A1A}
\definecolor{TPridgeBand}{HTML}{D88080}   
\definecolor{TPtest}     {HTML}{0E7C46}
\definecolor{TPpanelBg}  {HTML}{F7F8FB}
\definecolor{TPpanelEdge}{HTML}{DDE1E8}
\definecolor{TPtxtGray}  {HTML}{555555}
\definecolor{TParrow}    {HTML}{6B6B6B}
\definecolor{TPdistLine} {HTML}{0E7C46}

\pgfplotsset{
  panelaxis/.style={
    axis line style={draw=none},
    tick style={draw=none},
    xtick=\empty, ytick=\empty,
    enlargelimits=false,
    scale only axis=true,
    axis background/.style={fill=TPpanelBg, draw=TPpanelEdge, line width=0.4pt},
  },
  axislabeled/.style={
    panelaxis,
    every axis label/.append style={font=\scriptsize, text=TPtxtGray},
    xlabel style={yshift=2pt}, ylabel style={xshift=2pt},
  },
}

\providecommand{\PW}{2.95cm}
\providecommand{\PH}{2.95cm}
\providecommand{\HG}{0.80cm}

\resizebox{\linewidth}{!}{%
\begin{tikzpicture}[
    font=\rmfamily,
    annot/.style    ={font=\tiny, text=black, inner sep=1pt},
    annotmath/.style={font=\scriptsize, text=black, inner sep=1pt},
    caption/.style  ={font=\footnotesize\itshape, text=TPtxtGray, inner sep=0pt,
                      align=center},
    arrowlbl/.style ={font=\scriptsize, text=black, inner sep=1pt, align=center},
    arrowsub/.style ={font=\tiny\itshape, text=TPtxtGray, inner sep=1pt, align=center},
  ]

\node[anchor=north west, inner sep=0] (P1) at (0,0) {%
  \begin{tikzpicture}
    \begin{axis}[panelaxis,
                 width=\PW, height=\PH,
                 xmin=-1.10, xmax=1.10, ymin=-1.10, ymax=1.10]
      \addplot[TPcorrect, line width=0.9pt, smooth, samples=80, domain=-0.95:0.95]
        ({x},{0.55*x + 0.05*sin(deg(3*x))});
      \addplot[TPcorrect, line width=0.9pt, smooth, samples=80, domain=-0.95:0.85]
        ({x-0.05},{0.35*x - 0.20 + 0.04*sin(deg(2.5*x))});
      \addplot[TPcorrect, line width=0.9pt, smooth, samples=80, domain=-0.85:0.95]
        ({x+0.02},{0.45*x + 0.30 + 0.06*cos(deg(2.0*x))});
      \addplot[only marks, mark=*, mark size=0.5pt, color=TPcorrect]
        coordinates {(-0.8,-0.39)(-0.4,-0.20)(0.0,0.05)(0.4,0.27)(0.8,0.45)};
      \addplot[TPincorrect, line width=0.7pt, smooth, samples=120, domain=-0.85:0.85]
        ({x},{0.40*sin(deg(4*x+1.0)) - 0.10});
      \addplot[TPincorrect, line width=0.7pt, smooth, samples=120, domain=-0.85:0.85]
        ({x+0.05},{0.50*sin(deg(5*x)) + 0.30});
      \addplot[only marks, mark=*, mark size=0.9pt, color=TPcorrect]
        coordinates {(-0.95,-0.55)(0.95,0.50)};
    \end{axis}
  \end{tikzpicture}};

\node[annot, anchor=north west]
      at ($(P1.north west)+(0.04*\PW, -0.05*\PH)$)
  {\tikz[baseline=-0.5ex]{\draw[TPcorrect,line width=0.9pt] (0,0)--(6pt,0);}\,$y_i{=}1$\,
   \tikz[baseline=-0.5ex]{\draw[TPincorrect,line width=0.7pt] (0,0)--(6pt,0);}\,$y_i{=}0$};

\node[annot, anchor=south west]
      at ($(P1.south west)+(0.06*\PW, 0.04*\PH)$)
  {$\mathbf{H}_{i,j}\!\in\!\mathbb{R}^{T_{i,j}\!\times\!D}$};

\node[anchor=north west, inner sep=0]
      (P2) at ($(P1.north east)+(\HG,0)$) {%
  \begin{tikzpicture}
    \begin{axis}[axislabeled,
                 width=\PW, height=\PH,
                 xmin=-0.05, xmax=1.05, ymin=-0.05, ymax=1.40,
                 xlabel={\footnotesize $R^2$}]
      \addplot[only marks, mark=*, mark size=1.0pt, color=TPcorrect]
        coordinates {
          (0.92, 0.10)(0.88, 0.15)(0.95, 0.08)(0.85, 0.20)(0.82, 0.25)
          (0.78, 0.30)(0.74, 0.35)(0.70, 0.42)(0.66, 0.48)(0.62, 0.55)
          (0.58, 0.60)(0.54, 0.65)(0.50, 0.70)(0.46, 0.74)(0.42, 0.78)
          (0.38, 0.80)(0.34, 0.82)(0.30, 0.84)(0.90, 0.12)(0.83, 0.18)
          (0.75, 0.33)(0.68, 0.46)(0.60, 0.58)(0.52, 0.68)(0.44, 0.76)
          (0.36, 0.81)(0.80, 0.22)(0.72, 0.36)(0.64, 0.50)(0.56, 0.62)
          (0.48, 0.72)(0.40, 0.79)(0.28, 0.85)(0.32, 0.83)};
      \addplot[only marks, mark=o, mark size=1.1pt, color=TPincorrect,
               line width=0.35pt]
        coordinates {
          (0.15, 0.50)(0.20, 1.10)(0.25, 0.30)(0.10, 0.90)(0.30, 1.20)
          (0.45, 0.15)(0.55, 1.25)(0.65, 1.10)(0.18, 0.70)(0.35, 0.45)
          (0.50, 0.90)(0.70, 0.20)(0.40, 1.30)(0.22, 0.05)(0.60, 0.25)
          (0.05, 0.55)(0.78, 1.05)(0.08, 1.15)(0.42, 0.55)(0.85, 1.20)};
    \end{axis}
  \end{tikzpicture}};

\node[annot, anchor=north west]
      at ($(P2.north west)+(0.04*\PW, -0.05*\PH)$)
  {{\color{TPcorrect}$\bullet$}\,correct\,{\color{TPincorrect}$\circ$}\,incorrect};

\node[annot, anchor=west, font=\tiny\itshape, text=TPtxtGray]
      at ($(P2.west)+(0.04,0.30*\PH)$)
  {$K_n\!\uparrow$};

\node[annot, anchor=south west]
      at ($(P2.south west)+(0.08*\PW, 0.06*\PH)$)
  {$\mathcal{X}_{\text{traj}}\!\subset\!\mathbb{R}^{6}$};

\node[anchor=north west, inner sep=0]
      (P3) at ($(P2.north east)+(\HG,0)$) {%
  \begin{tikzpicture}
    \begin{axis}[axislabeled,
                 width=\PW, height=\PH,
                 xmin=-0.05, xmax=1.05, ymin=-0.05, ymax=1.40,
                 xlabel={\footnotesize $R^2$}]
      \addplot[fill=TPcorrect!10, draw=TPcorrect!22, line width=0.3pt,
               smooth cycle, tension=0.7]
        coordinates {
          (0.98,-0.02)(0.92,0.05)(0.80,0.20)(0.65,0.40)(0.48,0.62)
          (0.32,0.78)(0.22,0.88)(0.20,0.98)(0.28,1.00)(0.42,0.92)
          (0.58,0.78)(0.74,0.52)(0.88,0.28)(1.00,0.10)
        };
      \addplot[fill=TPcorrect!22, draw=TPcorrect!35, line width=0.3pt,
               smooth cycle, tension=0.7]
        coordinates {
          (0.96,0.02)(0.88,0.13)(0.76,0.28)(0.62,0.48)(0.46,0.68)
          (0.32,0.82)(0.28,0.90)(0.36,0.88)(0.50,0.75)(0.66,0.52)
          (0.82,0.27)(0.94,0.12)
        };
      \addplot[only marks, mark=*, mark size=1.0pt, color=TPcorrect]
        coordinates {
          (0.92, 0.10)(0.88, 0.15)(0.95, 0.08)(0.85, 0.20)(0.82, 0.25)
          (0.78, 0.30)(0.74, 0.35)(0.70, 0.42)(0.66, 0.48)(0.62, 0.55)
          (0.58, 0.60)(0.54, 0.65)(0.50, 0.70)(0.46, 0.74)(0.42, 0.78)
          (0.38, 0.80)(0.34, 0.82)(0.30, 0.84)(0.90, 0.12)(0.83, 0.18)
          (0.75, 0.33)(0.68, 0.46)(0.60, 0.58)(0.52, 0.68)(0.44, 0.76)
          (0.36, 0.81)(0.80, 0.22)(0.72, 0.36)(0.64, 0.50)(0.56, 0.62)
          (0.48, 0.72)(0.40, 0.79)(0.28, 0.85)(0.32, 0.83)};
      \addplot[TPridgeBand, line width=0.6pt, smooth, opacity=0.55] table[row sep=\\] {
        x  y \\  0.96 0.10 \\  0.88 0.18 \\  0.78 0.34 \\  0.66 0.51 \\
        0.54 0.66 \\  0.42 0.79 \\  0.30 0.86 \\
      };
      \addplot[TPridgeBand, line width=0.6pt, smooth, opacity=0.55] table[row sep=\\] {
        x  y \\  0.95 0.04 \\  0.86 0.14 \\  0.76 0.28 \\  0.64 0.47 \\
        0.52 0.62 \\  0.40 0.76 \\  0.28 0.84 \\
      };
      \addplot[TPridgeBand, line width=0.6pt, smooth, opacity=0.55] table[row sep=\\] {
        x  y \\  0.97 0.08 \\  0.90 0.16 \\  0.80 0.30 \\  0.68 0.44 \\
        0.56 0.60 \\  0.44 0.76 \\  0.32 0.82 \\
      };
      \addplot[TPridgeBand, line width=0.6pt, smooth, opacity=0.55] table[row sep=\\] {
        x  y \\  0.94 0.02 \\  0.84 0.12 \\  0.74 0.30 \\  0.60 0.52 \\
        0.48 0.70 \\  0.36 0.80 \\  0.26 0.86 \\
      };
      \addplot[TPridgeBand, line width=0.6pt, smooth, opacity=0.55] table[row sep=\\] {
        x  y \\  0.96 0.06 \\  0.88 0.20 \\  0.78 0.32 \\  0.66 0.46 \\
        0.54 0.64 \\  0.42 0.78 \\  0.30 0.85 \\
      };
    \end{axis}
  \end{tikzpicture}};

\node[annot, anchor=north west]
      at ($(P3.north west)+(0.04*\PW, -0.05*\PH)$)
  {{\color{TPcorrect}$\bullet$}\,$\mathcal{X}_{\text{traj}}^{+}$\,
   \tikz[baseline=-0.5ex]{\draw[TPridgeBand,line width=1.0pt] (0,0)--(6pt,0);}\,raw ridge};

\node[annot, anchor=south west]
      at ($(P3.south west)+(0.06*\PW, 0.06*\PH)$)
  {SCMS on $\hat p_h$ in $\mathbb{R}^{6}$};

\node[anchor=north west, inner sep=0]
      (P4) at ($(P3.north east)+(\HG,0)$) {%
  \begin{tikzpicture}
    \begin{axis}[axislabeled,
                 width=\PW, height=\PH,
                 xmin=-0.05, xmax=1.05, ymin=-0.05, ymax=1.40,
                 xlabel={\footnotesize $R^2$}]
      \addplot[only marks, mark=*, mark size=0.6pt, color=TPcorrect!40]
        coordinates {
          (0.92, 0.10)(0.88, 0.15)(0.85, 0.20)(0.78, 0.30)(0.70, 0.42)
          (0.62, 0.55)(0.54, 0.65)(0.46, 0.74)(0.38, 0.80)(0.30, 0.84)
          (0.80, 0.22)(0.72, 0.36)(0.64, 0.50)(0.56, 0.62)(0.48, 0.72)
          (0.40, 0.79)};
      \addplot[TPridge, line width=1.4pt, smooth] table[row sep=\\] {
        x  y \\
        0.96 0.06 \\  0.90 0.13 \\  0.84 0.21 \\  0.78 0.31 \\
        0.72 0.40 \\  0.66 0.49 \\  0.60 0.58 \\  0.54 0.66 \\
        0.48 0.73 \\  0.42 0.78 \\  0.36 0.82 \\  0.30 0.84 \\
        0.26 0.85 \\
      };
      \addplot[only marks, mark=square*, mark size=1.5pt, color=TPridgeV]
        coordinates {(0.94,0.075)(0.82,0.235)(0.70,0.425)(0.56,0.625)
                     (0.42,0.78)(0.30,0.84)};
    \end{axis}
  \end{tikzpicture}};

\node[annot, anchor=north west]
      at ($(P4.north west)+(0.04*\PW, -0.05*\PH)$)
  {\tikz[baseline=-0.5ex]{\draw[TPridge,line width=1.2pt] (0,0)--(6pt,0);}\,$\mathcal{R}_1$\,
   {\color{TPridgeV}$\blacksquare$}\,$\mathbf{v}_k$};

\node[annot, anchor=south west]
      at ($(P4.south west)+(0.06*\PW, 0.06*\PH)$)
  {clamp to $r{=}1$};

\node[anchor=north west, inner sep=0]
      (P5) at ($(P4.north east)+(\HG,0)$) {%
  \begin{tikzpicture}
    \begin{axis}[axislabeled,
                 width=\PW, height=\PH,
                 xmin=-0.05, xmax=1.05, ymin=-0.05, ymax=1.40,
                 xlabel={\footnotesize $R^2$}]
      \addplot[only marks, mark=*, mark size=0.6pt, color=TPcorrect!35]
        coordinates {
          (0.92, 0.10)(0.88, 0.15)(0.85, 0.20)(0.78, 0.30)(0.70, 0.42)
          (0.62, 0.55)(0.54, 0.65)(0.46, 0.74)(0.38, 0.80)(0.30, 0.84)};
      \addplot[TPridge, line width=1.2pt, smooth] table[row sep=\\] {
        x  y \\
        0.96 0.06 \\  0.90 0.13 \\  0.84 0.21 \\  0.78 0.31 \\
        0.72 0.40 \\  0.66 0.49 \\  0.60 0.58 \\  0.54 0.66 \\
        0.48 0.73 \\  0.42 0.78 \\  0.36 0.82 \\  0.30 0.84 \\
        0.26 0.85 \\
      };
      \addplot[only marks, mark=square*, mark size=1.3pt, color=TPridgeV]
        coordinates {(0.94,0.075)(0.82,0.235)(0.70,0.425)(0.56,0.625)
                     (0.42,0.78)(0.30,0.84)};
      \addplot[only marks, mark=triangle*, mark size=2.0pt, color=TPtest]
        coordinates {(0.74, 0.48)};
      \addplot[only marks, mark=triangle*, mark size=2.0pt, color=TPtest]
        coordinates {(0.55, 0.85)};
      \addplot[only marks, mark=triangle*, mark size=2.0pt, color=TPtest]
        coordinates {(0.20, 0.20)};
      \addplot[TPdistLine, line width=0.5pt, dashed, dash pattern=on 1.2pt off 0.9pt]
        coordinates {(0.74, 0.48) (0.70, 0.425)};
      \addplot[TPdistLine, line width=0.5pt, dashed, dash pattern=on 1.2pt off 0.9pt]
        coordinates {(0.55, 0.85) (0.48, 0.73)};
      \addplot[TPdistLine, line width=0.5pt, dashed, dash pattern=on 1.2pt off 0.9pt]
        coordinates {(0.20, 0.20) (0.30, 0.84)};
    \end{axis}
  \end{tikzpicture}};

\node[annot, anchor=north west]
      at ($(P5.north west)+(0.04*\PW, -0.05*\PH)$)
  {{\color{TPtest}$\blacktriangle$}\,$\mathbf{x}_{\star,j}$\,
   {\color{TPdistLine}- -}\,$z_{\text{off}}$};

\node[annot, anchor=south east]
      at ($(P5.south east)+(-0.04*\PW, 0.05*\PH)$)
  {$s(q_\star)\!=\!-\overline{z_{\text{off}}}$};

\foreach \L/\R/\toplbl/\botlbl in {%
  P1/P2/{$\varphi:\mathbb{R}^{T\!\times\!D}\!\to\!\mathbb{R}^{6}$}/{feature map},%
  P2/P3/{restrict $y{=}1$}/{SCMS in $\mathbb{R}^{6}$},%
  P3/P4/{$\pi$}/{est.\ dim., clamp $r{=}1$},%
  P4/P5/{$\min_k\!\|\cdot\!-\!\mathbf{v}_k\|$}/{nearest-vertex}%
}{
  \coordinate (aL) at ($(\L.north east)!0.5!(\L.south east)+(0.06,0)$);
  \coordinate (aR) at ($(\R.north west)!0.5!(\R.south west)+(-0.06,0)$);
  \path let \p1=(aL), \p2=(aR) in coordinate (aRh) at (\x2,\y1);
  \draw[-{Stealth[length=2.2mm,width=1.8mm]}, TParrow, line width=0.75pt]
    (aL) -- (aRh);
  \node[arrowlbl, anchor=south] at ($(aL)!0.5!(aRh)+(0,0.05)$) {\toplbl};
  \node[arrowsub, anchor=north] at ($(aL)!0.5!(aRh)+(0,-0.05)$) {\botlbl};
}

\node[caption, anchor=north] at ($(P1.south)+(0,-0.18)$)
  {\textbf{(a)} trajectories};
\node[caption, anchor=north] at ($(P2.south)+(0,-0.18)$)
  {\textbf{(b)} 6-D feature cloud};
\node[caption, anchor=north] at ($(P3.south)+(0,-0.18)$)
  {\textbf{(c)} raw SCMS ridge};
\node[caption, anchor=north] at ($(P4.south)+(0,-0.18)$)
  {\textbf{(d)} 1-D ridge $\mathcal{R}_1$};
\node[caption, anchor=north] at ($(P5.south)+(0,-0.18)$)
  {\textbf{(e)} score $s(q_\star)$};

\end{tikzpicture}%
}
}

%% file: tables/ridge_variants.tex
\begin{table*}[t]
\centering
\caption{Ridge embedding variants. All three start from the same SCMS density-ridge estimate and differ in how the ridge is converted into coordinates.}
\label{tab:ridgevariants}
\small
\begin{tabular}{llll}
\toprule
Method & Constructs & Output Dimension & Captures \\
\midrule
Ridege LTSA Chart & 1 global LTSA & $r$ & Global tangent structure \\
Ridge Arclength  & 1-D geodesic on ridge graph & $r{+}1$ (arclen $+$ $z_{off}$) & Progress along a curve \\
Ridge Atlas      & Many local charts stitched  & $r{+}1$ (local $+$ $z_{off}$)  & Curved / multi-patch manifolds \\
\bottomrule
\end{tabular}
\end{table*}

%% file: tables/selective_pred_table.tex
\begin{table}[!htbp]
    \centering
\caption{Selective prediction comparison on select methods ($q=200, N=5, \texttt{test size} = 60$). Ridge estimation methods used on hidden state trajectory sequences are compared against calibrated log probability, and a non-ridge method. Arrows $\uparrow$/$\downarrow$ indicate higher/lower is better. Most performant predictor per (model, dataset) is in \textbf{bold}.}
\label{tab:selective_pred_combined}
\scriptsize
\setlength{\tabcolsep}{1.2pt}
\renewcommand{\arraystretch}{1.15}
\begin{tabular*}{\columnwidth}{@{\extracolsep{\fill}} p{1.6cm} p{1.3cm} >{\tiny}p{2.6cm} c c c c @{}}
\toprule
\textbf{Model} & \textbf{Dataset} & \multicolumn{1}{l}{\textbf{Method (Varient/Scorer) }} &
\multicolumn{1}{c}{\tiny AUROC$\uparrow$} &
\multicolumn{1}{c}{\tiny AURC$\downarrow$} &
\multicolumn{1}{c}{\tiny PRR$\uparrow$} &
\multicolumn{1}{c}{\tiny AUGRC$\downarrow$} \\
\midrule
    Idefics3-8B-Llama3 & A-OKVQA & logP(x) \textit{(Sequence logP(x))} & 0.795 & 0.244 & 0.752 & 0.152 \\
     &  & Topology LID-MLE \textit{(Neg LID-MLE)} & 0.725 & 0.252 & 0.744 & 0.169 \\
     &  & Ridge LTSA Chart Baseline \textit{(PR-dim/Ridge Proximity)} & \textbf{0.972} & \textbf{0.132} & \textbf{0.864} & \textbf{0.108} \\
    \midrule

    SmolVLM-Instruct & A-OKVQA & Baseline logP(x) \textit{(Sequence logP(x))} & 0.791 & 0.344 & 0.651 & 0.221 \\
     &  & Topology LID-MLE \textit{(Neg LID-MLE)} & 0.470 & 0.593 & 0.402 & 0.299 \\
     & & Ridge LTSA Chart \textit{(PR-dim/Ridge Proximity)} & \textbf{0.951} & \textbf{0.245} & \textbf{0.750} & \textbf{0.182} \\
    \midrule
    
    Idefics3-8B-Llama3 & ScienceQA &  Baseline kNN \textit{(kNN-R² proximity)} & 0.907 & 0.163 & 0.833 & 0.117 \\
     &  & Topology LID-MLE \textit{(Neg LID-MLE)} & 0.511 & 0.526 & 0.479 & 0.214 \\
     && Ridge Arclength \textit{(Spherized/Ridge Proximity)} & \textbf{0.934} & \textbf{0.140} & \textbf{0.856} & \textbf{0.110} \\
    \midrule

    SmolVLM-Instruct & ScienceQA &  Baseline logP(x) \textit{(Sequence logP(x))} & 0.790 & 0.243 & 0.753 & 0.169 \\
     &  & Flow Matching \textit{(Neg dist. to correct centroid)} & 0.769 & 0.326 & 0.670 & 0.175 \\
     & & Ridge Arclength \textit{(PR-dim/Ridge Proximity)} & \textbf{0.990} & \textbf{0.146} & \textbf{0.850} & \textbf{0.119} \\
    \hline\hline \\

    Mistral-7B-\linebreak Instruct-v0.3 & HaluEval-QA & Baseline logP(x) \textit{(Sequence logP(x))} & 0.913 & 0.202 & 0.794 & 0.147 \\
     &  & Traced Mean Curvature & 0.892 & 0.203 & 0.793 & 0.152 \\
     & & Ridge Arclength \textit{(Shrinkage/Ridge Proximity)} & \textbf{0.971} & \textbf{0.169} & \textbf{0.826} & \textbf{0.132} \\
    \midrule
    
    LLaVA-v1.5-7B & POPE & Baseline logP(x) \textit{(Mean logP(x) per token)} & 0.817 & 0.106 & 0.891 & 0.077 \\
     &  & Semantic Entropy & 0.740 & 0.133 & 0.864 & 0.093 \\
     & & Ridge LTSA Chart  \textit{(TRiE/Ridge Proximity)} & \textbf{1.000} & \textbf{0.044} & \textbf{0.953} & \textbf{0.040} \\
    \midrule
    
    Gemma-2-9B-IT & GSM8K  & Baseline kNN \textit{(kNN-R² proximity)} & 0.813 & 0.012 & 0.988 & 0.010 \\
     &  & Neg. H0 total persistence & 0.637 & 0.024 & 0.975 & 0.018 \\
     & & Ridge Atlas \textit{(TRiE/Ridge Proximity)} & \textbf{0.994} & \textbf{0.002} & \textbf{0.998} & \textbf{0.002} \\
    \bottomrule
    \end{tabular*}
    \vspace{2pt}
    \begin{flushleft}\tiny
        {\tiny
        \textbf{Method legend.} All methods, except the log probability baselines, produce a per-query scalar confidence score, where higher confidence can be used as a selective prediction signal; scores are negated to standardized orientation to higher being more confident. 
        Trajectories are acquired from sequences of hidden state final layer last tokens across $T$ autoregressive decoding steps (shape $(T, D)$), aggregated across $N$ sampled generations.
        \emph{Ridge-based (ours):}
        \textbf{Ridge Arclength}, \textbf{Ridge LTSA Chart}, \textbf{Ridge Atlas},described in detail in Table \ref{tab:ridgevariants}, fit a subspace constrained mean shift (SCMS) principal ridge on a projection of the hidden states of the training set (correct-only subset); 
        the score is the perpendicular distance $\mathrm{z_{off}}$ from the test query's hidden state to its nearest ridge vertex (\emph{Ridge Proximity}). 
        Ablated variants of SCMS variants are utilized to ascertain the effect of ridge projection and density estimation on trajectory-computed selective prediction; ablation variants in the table include \emph{Shrinkage} (covariance ablation), \emph{Trajectory Ridge Estimate (TRiE)} (no ablation), \emph{Spherized} (projection ablation via per-row $L2$ normalization instead of identity projection), and \emph{PR-dim} (projection and covariance ablation).
        \emph{Baselines:}
        \textbf{Baseline logP(x)} — \emph{Sequence}: sum of token log-probs; \emph{Mean/token}: length-normalized.
        \textbf{Baseline kNN (kNN-$R^2$ proximity)} — local $R^2$-style statistic over the labeled training set near the query's hidden state.
        \textbf{Semantic Entropy} \cite{kuhn2023} — NLI-clusters $N$ generations into semantic equivalence classes, Shannon entropy over cluster probabilities (negated for confidence).
        \textbf{Flow Matching} — train a CFM vector field on correct-class hidden states (PCA to TwoNN dim); integrate ODE backward to obtain the base-Gaussian latent $z$; score $=$ negative distance from $z$ to correct-class centroid.
        \textbf{Topology LID-MLE} — negated MLE (maximum likelihood estimate) of local intrinsic dimension (LID) at the query; query's local hidden-state neighborhood.
        \textbf{Traced Mean Curvature} — trace of the mean-curvature tensor along each $(T, D)$ generation trajectory, aggregated across the $N$ generations \cite{trace2025}.
        \textbf{Neg.\ H$_0$ total persistence} — sum of bar lengths in the H$_0$ persistence diagram of the per-query hidden state under a Vietoris--Rips filtration (negated).
        \par}
    \end{flushleft}
    \end{table}

%% file: figures/metrics.tex
\begin{figure}[!htbp]
  \centering
  \begin{subfigure}[b]{\linewidth}
    \centering
    \includegraphics[width=0.75\linewidth]{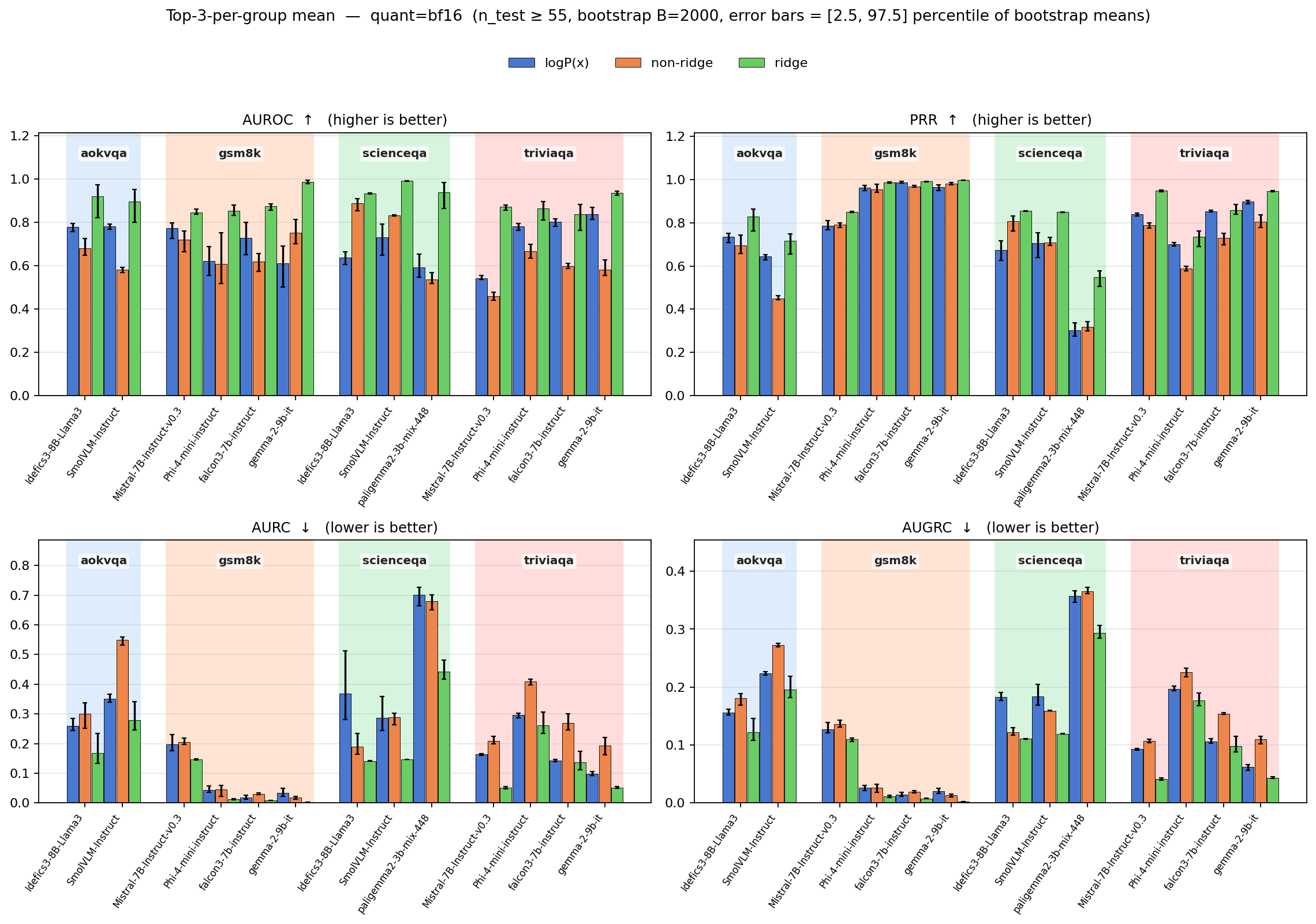}
    \caption{bf16 precision.}
    \label{fig:metrics-bf16}
  \end{subfigure}

  \vspace{0.5em}

  \begin{subfigure}[b]{\linewidth}
    \centering
    \includegraphics[width=0.75\linewidth]{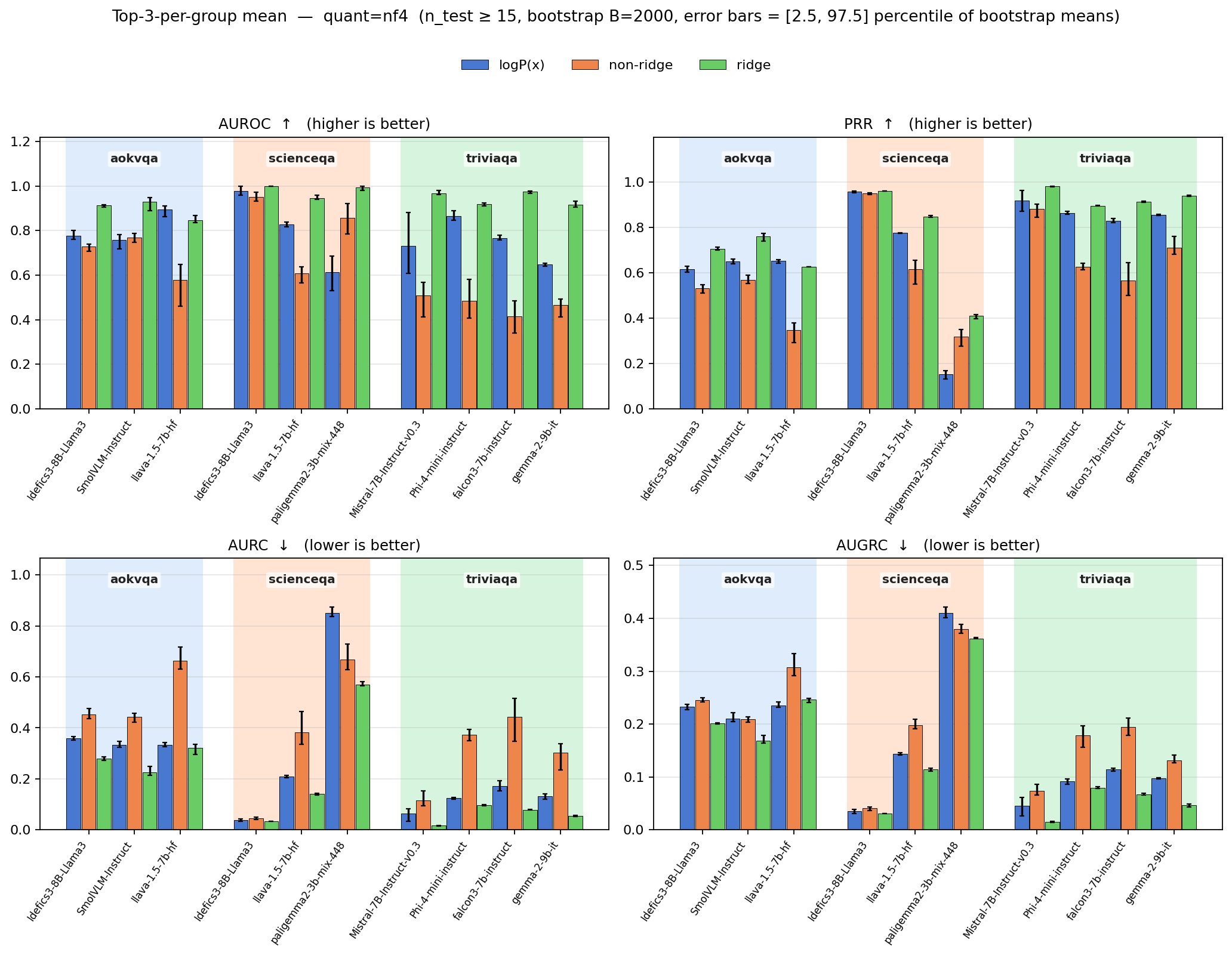}
    \caption{nf4 quantization.}
    \label{fig:metrics-nf4}
  \end{subfigure}
  \caption{Top-3 detectors across (model, dataset) cells, evaluated under
  bf16 precision (top) and nf4 quantization (bottom). Panels show AUROC
  and PRR (higher is better) and AURC and AUGRC (lower is better).}
  \label{fig:metrics-top3}
\end{figure}